\documentclass{article}
\usepackage{spconf,amsmath,graphicx}
\usepackage{multirow}
\usepackage{adjustbox}

\title{Image Deraining via Self-supervised Reinforcement Learning}

\name{He-Hao Liao$^{1}$, Yan-Tsung Peng$^{1}$, Wen-Tao Chu$^{2}$, Ping-Chun Hsieh$^{2}$, and Chung-Chi Tsai$^{3}$}
\address{$^{1}$ Dept. of Computer Science, National Chengchi University, Taiwan, \\$^{2}$ Dept. of Computer Science, National Yang Ming Chiao Tung University, Taiwan, \\ $^{3}$ Qualcomm Technologies, USA}

\begin{document}
\topmargin=0mm
%
\maketitle
\begin{abstract}
The quality of images captured outdoors is often affected by the weather. One factor that interferes with sight is rain, which can obstruct the view of observers and computer vision applications that rely on those images. The work aims to recover rain images by removing rain streaks via Self-supervised Reinforcement Learning (RL) for image deraining (SRL-Derain). We locate rain streak pixels from the input rain image via dictionary learning and use pixel-wise RL agents to take multiple inpainting actions to remove rain progressively. To our knowledge, this work is the first attempt where self-supervised RL is applied to image deraining. Experimental results on several benchmark image-deraining datasets show that the proposed SRL-Derain performs favorably against state-of-the-art few-shot and self-supervised deraining and denoising methods.
\end{abstract}
\section{Introduction}
\label{sec:intro}

Photos captured outside by personal cameras, dash-cams, surveillance cameras, and other devices can potentially capture scenes of rain, which may obscure the visibility of the images. This can decrease the effectiveness of subsequent computer vision applications, such as object detection and recognition. Therefore, it is essential to have a good method to remove rain from these images so that they can be used effectively.

Image deraining has been much researched. For example, traditional statistical methods separate rain streaks from an image of rain and obtain a clean background~\cite{li2016rain,decompositionSID_Kang2011TIP,wang2017hierarchical}. Li \emph{et al.}~\cite{li2016rain} utilized a priori information to extract rain-related characteristics and applied a Gaussian mixture model to distinguish between background and rain layers. Kang \emph{et al.}~\cite{decompositionSID_Kang2011TIP} applied the dictionary learning, k-means clustering, and sparse coding to separate the non-rain and rain components. Wang \emph{et al.}~\cite{wang2017hierarchical} used dictionary learning to separate rain or snow from the input image. However, these methods rely solely on statistical image priors to restore rain images without much guidance, often resulting in unsatisfying deraining results.

Low-level vision tasks have witnessed considerable progress thanks to deep-learning-based methods. In deraining, supervised approaches, leveraging a large amount of rain and rain-free paired training data, allow models to learn rain streaks across various scenes to remove them from rain images. However, collecting paired real-world rain and rain-free images is challenging. As a result, most current methods are trained on synthetic datasets. The domain gap between training and testing sets often leads to suboptimal performance on unseen data or real-world rain images. Wei \emph{et al.}~\cite{wei2021deraincyclegan} used unpaired images and a generative adversarial network (GAN) with unsupervised learning, but GAN-based models may introduce unexpected artifacts in the results. 

\begin{figure}[!t]
\centering
\includegraphics[width=\columnwidth]{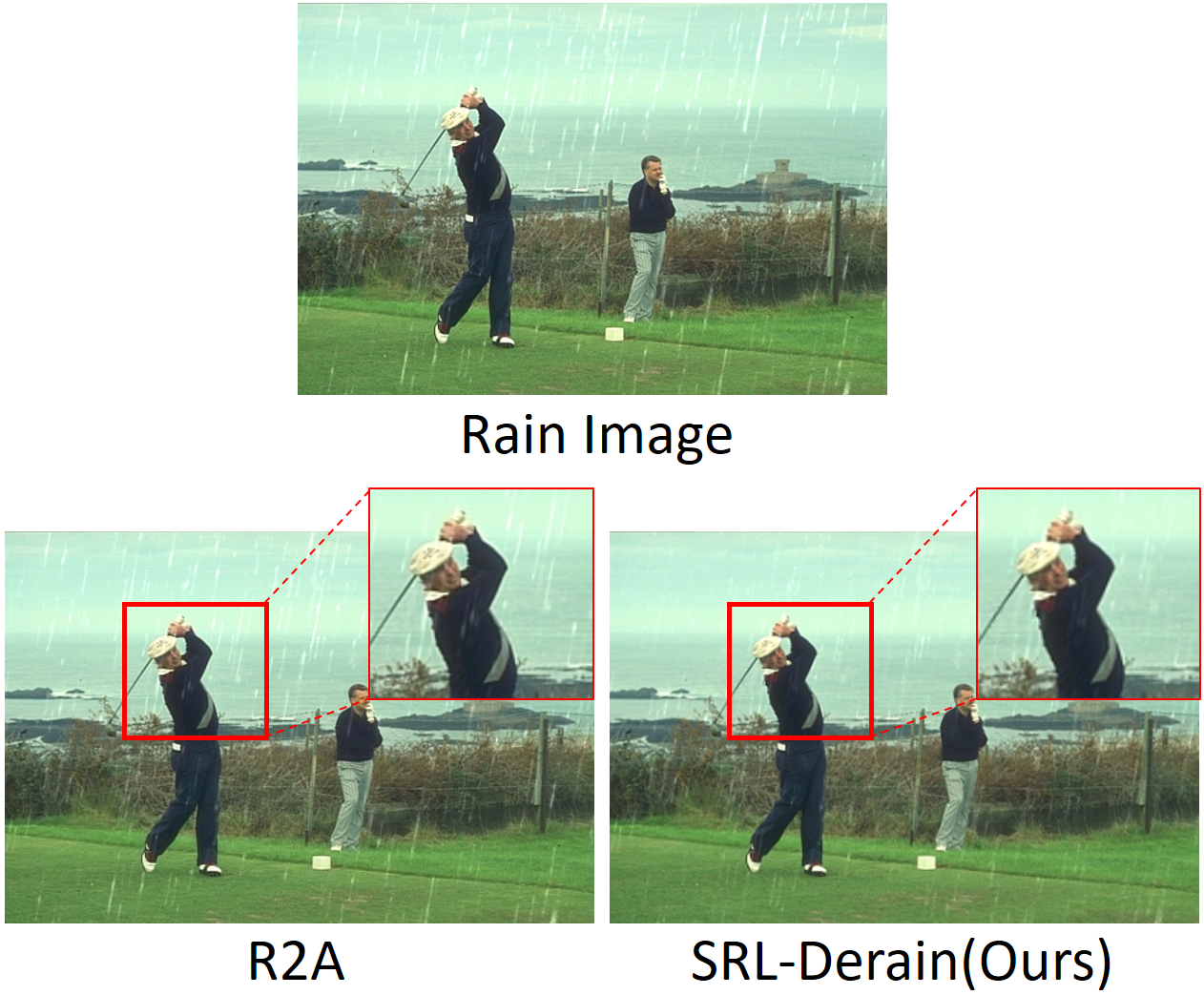}
\vspace{-2em}
\caption{A comparison of deraining results between R2A~\cite{peng2023rain2avoid}, a self-supervised learning method, and SRL-Derain, the proposed self-supervised reinforcement-learning method.}
\label{fig:ssl_srl_compare}
\vspace{-2em}
\end{figure}

The self-supervised deraining task has gained attention recently~\cite{peng2023rain2avoid}. However, only a few self-supervised methods have been developed specifically for deraining. Peng \emph{et al.}~\cite{peng2023rain2avoid} introduced a self-supervised deraining framework, Rain2Avoid (R2A). They proposed a locally dominant gradient prior (LDGP) to find rain pixels and avoided them upon training with the input rain image. However, its deraining performance is limited by its restoration ability from convolutional neural networks. This paper introduces SRL-Derain, a self-supervised RL approach for image deraining. We adopt dictionary learning as in~\cite{decompositionSID_Kang2011TIP} to locate rain streak pixels and pixel-wise RL agents~\cite{furuta2019fully} to take multiple inpainting actions to remove rain streaks from the input image gradually. The reward is calculated based on the pseudo-derained reference~\cite{peng2023rain2avoid} and BRISQUE~\cite{mittal2012no} score, a no-reference image quality metric that can be used to evaluate derained results. Figure~\ref{fig:ssl_srl_compare} compares the deraining results of R2A~\cite{peng2023rain2avoid} and SRL-Derain. Our self-supervised reinforcement-learning approach removes rain streaks better.

The contributions of this paper are threefold.
\begin{itemize}
    \item To our best knowledge, this is the first attempt at self-supervised reinforcement learning applied to image deraining.
    \item We use pseudo-deraining references and a no-reference image quality metric as self-supervised rewards to guide our proposed RL scheme training. 
    \item The experimental results show that the proposed SRL-Derain outperforms state-of-the-art few-shot and self-supervised methods for deraining and denoising.
\end{itemize}

\section{RELATED WORK}
\label{sec:related}

\subsection{Self-supervised Image Restoration}

Although few self-supervised methods have been proposed for image deraining, self-supervised denoising has been well-studied~\cite{ulyanov2018deep, krull2019noise2void, batson2019noise2self}. Ulyanov~\emph{et al.} found that the model tended to learn non-noise features more easily during training and proposed the Deep Image Prior (DIP) ~\cite{ulyanov2018deep}, in which random noise is fed into the neural network supervised with the noisy input image to obtain a denoised image. However, rain features differ from noise, making applying this approach to rain images impractical. N2V~\cite{krull2019noise2void} and N2S~\cite{batson2019noise2self} assume that noise in a noisy image is zero-mean and the noise between different pixels is independent. Based on the two assumptions, the network learns the denoised image with averaged noise pixels. Yet, these two assumptions do not apply to rain images and thus cannot effectively remove rain streaks. Peng~\emph{et al.} ~\cite{peng2023rain2avoid} introduced a self-supervised deraining scheme. They predicted the location of rain streaks based on the Locally Dominant Gradient Prior (LDGP) and generated pseudo-derained references for deraining. However, the self-supervised learning scheme using a pure convolutional neural network to learn from imperfect reference images has limitations in restoring derained images. We discovered that RL equipped with a set of conventional filtering tools is a superior method in the Self-Supervised Learning (SSL) regime for image restoration, where only limited information can be provided from a single image.

\subsection{Reinforcement-Learning-Based Image Restoration}
After the success of Deep Q Learning in RL, it has been broadly applied to tasks such as video game playing, robotics, and autonomous driving. Recently, researchers have further applied RL to solve computer vision tasks. Yu \emph{et al.}~\cite{yu2018crafting} addressed three types of image degradation: Gaussian blur, Gaussian noise, and JPEG compression. Park \emph{et al.} ~\cite{park2018distort} uses DQN to train a color enhancement network, which retouches the input image by manipulating the contrast, brightness, or color saturation iteratively. Additionally, Ryosuke \emph{et al.}~\cite{furuta2019fully} proposed a multi-agent image restoration framework to solve image denoising, image inpainting, and local color enhancement tasks.

Motivated by the success of SSL, various attempts have augmented RL methods with self-supervision. One major approach is to construct auxiliary tasks based on self-supervised prediction, which typically involves the prediction of forward transition dynamics or inverse dynamics, and use the prediction loss as an additional reward signal for improving the sample efficiency of RL~\cite{shelhamer2017loss}.
Another category leverages self-supervised learning to learn informative representations and thereby facilitate the downstream RL, especially for control problems with image-based observations. 
The principle behind this approach lies in the widely observed empirical evidence that RL from physical state features enjoys better sample efficiency than RL from raw images.
For example,~\cite{laskin2020curl} proposes CURL, which leverages contrastive learning to extract physical features from raw pixels through joint optimization of the self-supervised and RL losses. This joint optimization framework has also been instantiated through either reconstruction or self-prediction~\cite{schwarzer2021data}.
In addition to joint optimization, self-supervised pre-training has also shown good promise in boosting the data efficiency of downstream RL~\cite{higgins2017darla}.
Despite the above progress on SSL-augmented RL, one rather underexplored research question is how RL could benefit self-supervised vision tasks.

\section{METHODOLOGY}
\label{sec:method}

\begin{figure*}[!t]
\centering
\includegraphics[width=2\columnwidth]{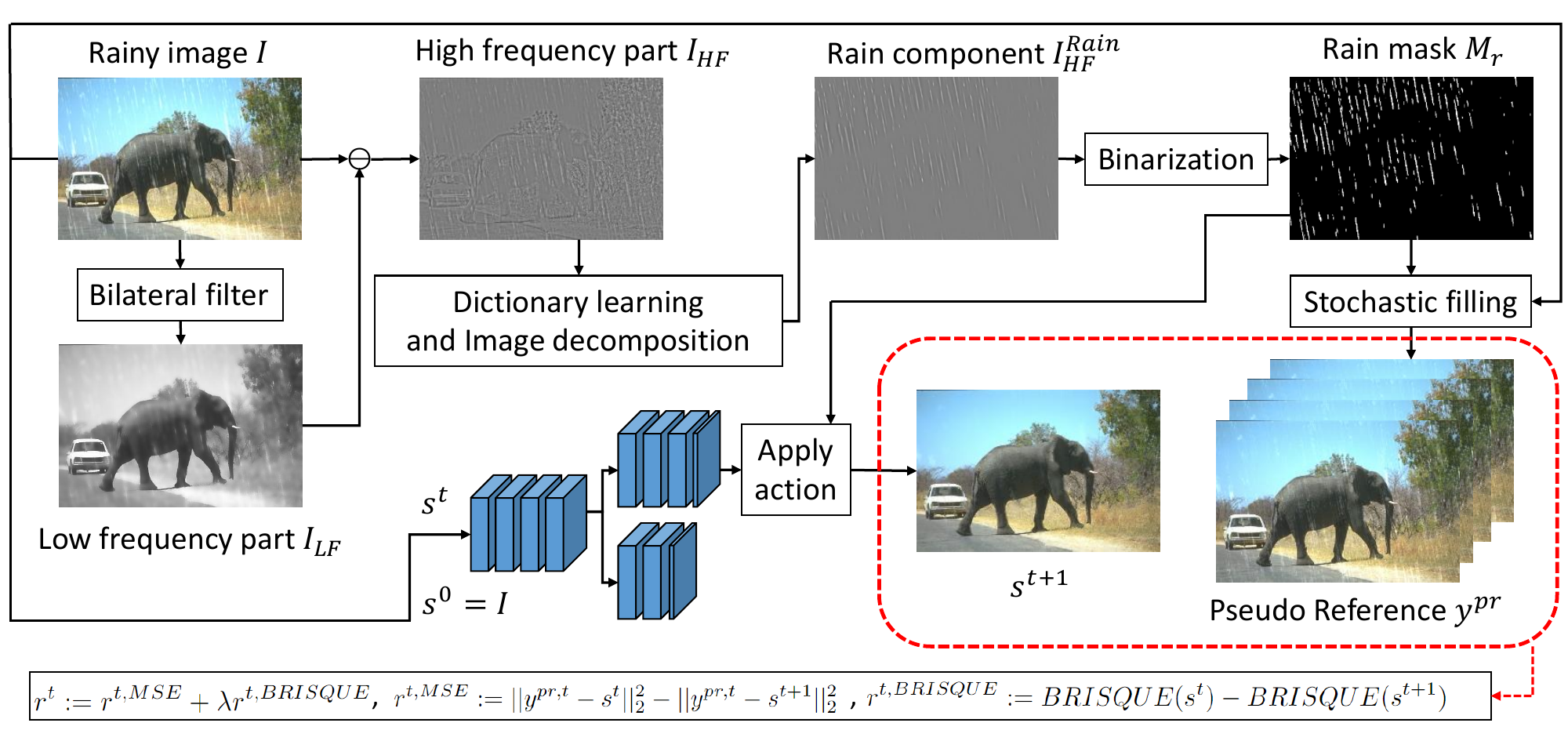}
\vspace{-1em}
\caption{The flowchart of the proposed self-supervised RL-based deraining scheme. To locate rain pixels and generate the rain mask, we utilize bilateral filtering to extract the high-frequency part $I_{HF}$ from the input rain image and decompose the rain components via dictionary learning~\cite{decompositionSID_Kang2011TIP}. The RL model progressively fills the rain pixels in the rain image based on the rain mask. Note that $r^{t} \in R^{H\times W}$ is the total reward map for the state $s^t$ at the time step $t$. }
\label{fig:arch}
\end{figure*}

This section details the proposed SRL-Derain. We will start by explaining how we extract rain streaks from the input image and generate pseudo-derained references. After that, we will introduce the multi-agent RL model, which is trained using pseudo-derained references and a no-reference image quality metric. Figure~\ref{fig:arch} depicts the overall flowchart of the proposed SRL-Derain.

\subsection{Generation of the Rain Mask and Pseudo-derained References}
\noindent\textbf{Rain Mask Generation.} There are several methods~\cite{jiang2018fastderain, peng2023rain2avoid} to separate rain streaks from rain images. Jiang \emph{et al.}~\cite{jiang2018fastderain} calculated the discriminatively intrinsic prior using the unidirectional total variation of the rain direction, determined based on the differences between consecutive video frames. However, it is not suitable for single-image deraining. Peng \emph{et al.}~\cite{peng2023rain2avoid} proposed Locally Dominant Gradient Prior (LDGP), calculating the Histogram of Oriented Gradients (HoGs) for each local region of a rain image and performing a majority vote to determine the rain-streak angle. After that, they extracted the rain streaks based on the angle using image morphological operations. However, it can only find the rain streaks at one angle. 
By contrast, we adopt dictionary learning as in ~\cite{decompositionSID_Kang2011TIP} to locate rain pixels, which we call Rain Dictionary Prior (RDP). First, using a bilateral filter, we decompose the rain image into low- and high-frequency parts $I_{LF}$ and $I_{HF}$. Next, we construct a dictionary $D\in r^{n\times m}$ for $I_{HF}$ using online dictionary learning~\cite{mairal2010online} as:
\begin{equation}
    \mathbf{argmin}_{D, \theta^i_d} \frac{1}{N_p} \sum^{N_p}_{i=1} \Big( \frac{1}{2} \|y^i_{HF}-D\theta^i_d\|_2^2+\lambda_d\|\theta^i_d\|_1 \Big),
\end{equation}
where $y^i_{HF} \in R^n$ is $p\times p$ overlapped patches centered at each pixel obtained from $I_{HF}$ and vectorized with size $n=p^2$, $N_p$ is the number of total patches extracted, $\theta^i_d$ is the sparse coefficients for the dictionary $D$ to construct $y^i_{HF}$, and $m$ is the number of atoms in the dictionary. Next, to extract rain-related atoms from the dictionary, each $y^i_{HF}$ is described by the HOG descriptor and classified into two clusters by the K-means algorithm. The cluster with the lower variance represents the rain-related atoms since rain streaks are geometrically simple, as suggested in~\cite{decompositionSID_Kang2011TIP}. Then, we utilize the rain-related atoms in the dictionary to reconstruct the rain component $I^{Rain}_{HF}$, followed by binarization to obtain the rain mask $M_r$ as the RDP.

\vspace{1em}

\noindent\textbf{Pseudo-derained Reference Generation.} To train our multi-agent RL model in a self-supervised manner, we use the rain mask generated by R2A~\cite{peng2023rain2avoid} to create pseudo-derained references. We then identify rain pixels from the rain mask and replace them with non-rain pixels around them in a stochastic manner. This means that for each rain pixel indicated by the computed RDP, a non-rain pixel in its neighborhood is randomly selected to generate a pseudo-derained reference $y^{pr}$ for the RL's reward function later. 

\begin{figure*}[t!]
\centering
\includegraphics[width=2\columnwidth]{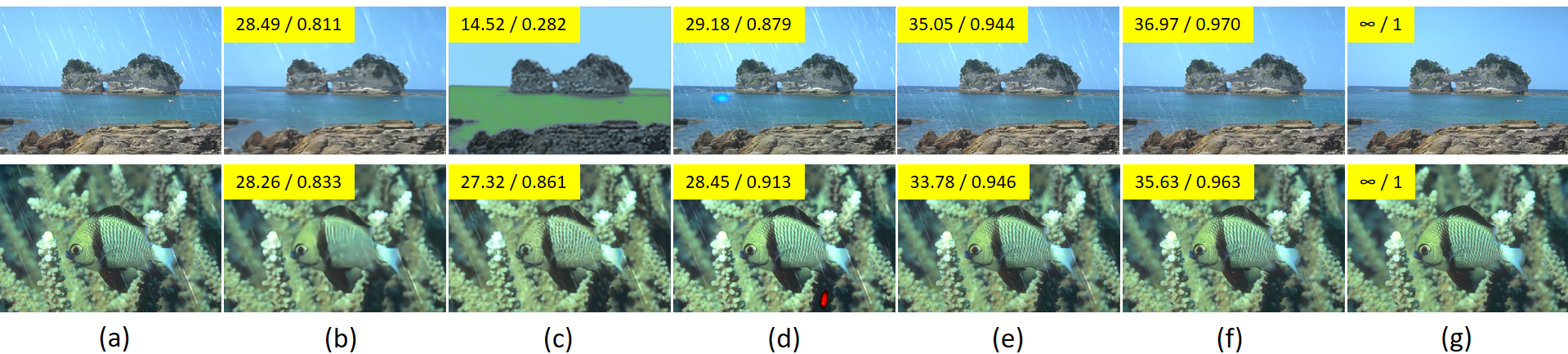}
\vspace{-1em}
\caption{Qualitative comparisons on Rain100L with PSNR/SSIM values shown on the results. (a) Input images, and the derained results obtained using (b) DIP~\cite{ulyanov2018deep}, (c) N2S~\cite{batson2019noise2self}, (d) N2V~\cite{krull2019noise2void}, (e) R2A~\cite{peng2023rain2avoid}, and (f) Ours. (g) GT images.}
\label{fig: Quality_rain100L}
\vspace{-0.5em}
\end{figure*}

\begin{figure*}[t!]
\centering
\includegraphics[width=2\columnwidth]{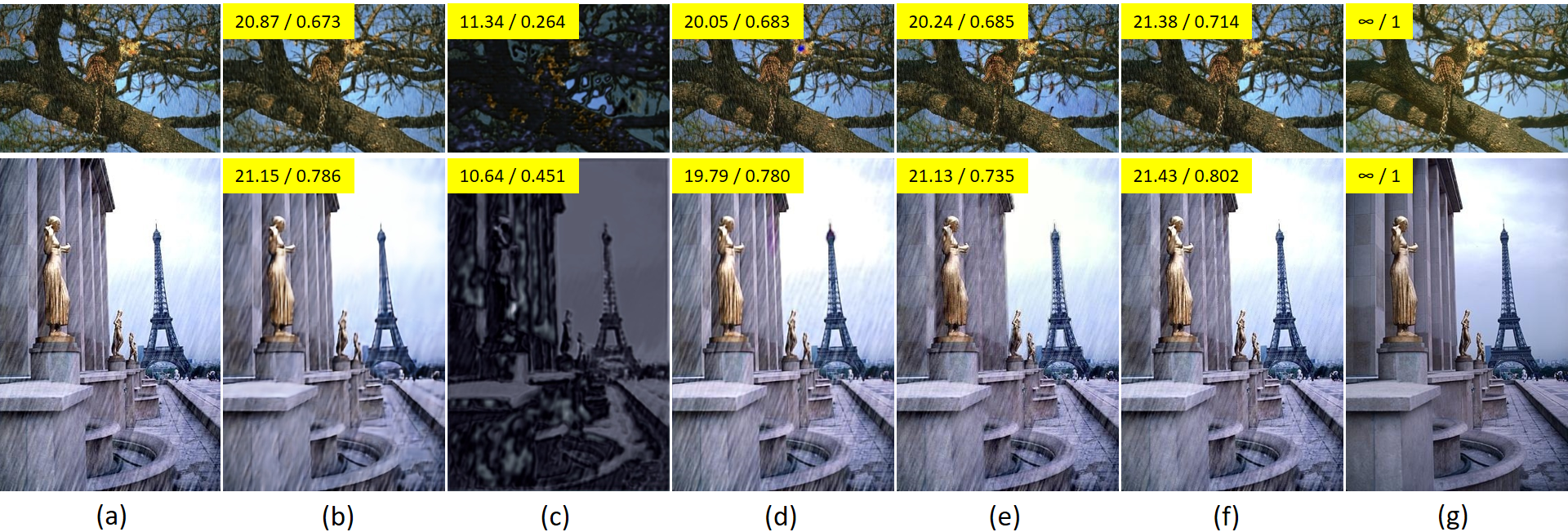}
\vspace{-1em}
\caption{Qualitative comparisons on Rain800 with PSNR/SSIM values shown on the results. (a) Input images, and the derained results obtained using (b) DIP~\cite{ulyanov2018deep}, (c) N2S~\cite{batson2019noise2self}, (d) N2V~\cite{krull2019noise2void}, (e) R2A~\cite{peng2023rain2avoid}, and (f) Ours. (g) GT images.}
\label{fig: Quality_rain800}
\vspace{-1em}
\end{figure*}

\subsection{RL-based Self-supervised Deraining Scheme}

We propose to cast image deraining as an RL problem with self-supervised rewards.
Specifically: (i) Regarding the RL formulation, we adopt the pixel-wise control as in pixelRL~\cite{furuta2019fully}, an RL-based image enhancement framework with one agent for each pixel based on the Asynchronous Advantage Actor-Critic (A3C) algorithm, to remove rain streaks from the input rain image, where the rain pixels are progressively restored with a clean background. 
Let $s_i^t$, $a_i^t$, and $r_i^t$ denote the image state, the action, and the self-supervised reward of the $i$-th pixel at each step $t$ of the an RL-based deraining episode, respectively. Let $\gamma\in [0,1)$ be the discount factor.
Built on the A3C method, SRL-Derain consists of a policy network $\pi_i(\cdot \rvert s; \theta_p)$ and a value network $V(\cdot \rvert s; \theta_v)$, which are parameterized by $\theta_p$ and $\theta_v$, respectively.
Let $\pi_i(\cdot \rvert s; \theta_p)$ denote the action distribution of the $i$-th pixel at state $s$ under the policy parameter $\theta_p$. 
The policy and value networks are iteratively updated by taking gradient steps of the corresponding loss functions defined as follows. 
In each iteration, we collect the trajectory of each pixel $\{s_i^1,a_i^1,r_i^1,\cdots, s_i^T,a_i^T,r_i^T\}$ and compute the value loss and the policy loss as
\begin{align}
    L_v(\theta_v)&:=\sum_{t=1}^T\frac{1}{N}\sum_{i=1}^N(R^{t}_{i}-V(s^{t}_{i};{\theta_v}))^2,\\
    L_p(\theta_p)&:=-\sum_{t=1}^T\sum_{i=1}^{N}\log \pi(a^{t}_{i}\rvert s^{t}_{i};\theta_p)A(a^{t}_{i}, s^{t}_{i}),
\end{align}
where $R_i^t$ and $A(a^{t}_{i}, s^{t}_{i})$ denotes the $n$-step return and the $n$-step return with baseline defined as
\begin{align}
    &R^{t}_{i}:= r^{t}_{i} + \gamma r^{t+1}_{i} + \gamma^{2}r^{t+2}_{i} + \cdots + \gamma^{n}V(s^{t+n}_{i};{\theta_v}),\\
    &A(a^{t}_{i}, s^{t}_{i}):=R^{t}_{i}-V(s^{t}_{i};{\theta_v}).
\end{align}
Following~\cite{furuta2019fully}, we can simultaneously train multiple pixel-wise agents using convolution neural networks.

To guide the training of the RL model, the proposed self-supervised rewards include a conventional mean-square-error-based difference between the state of the $i$-th agent (the $i$-th pixel in the intermediate derained image) and the corresponding pixel in the pseudo-derained reference $y^{pr,t}_i$ at the time step $t$, denoted as $r^{t,MSE}$ and the proposed no-reference quality metric reward, $r^{t,BRISQUE}$, described as follows:
\begin{align}
&r^{t, MSE}_{i}:= ||{{y}^{pr,t}_{i}-s^{t}_{i}}||^{2}_{2} - ||{y^{pr,t}_{i}-s^{t+1}_{i}}||^{2}_{2};\\
&r^{t, BRISQUE}_{i} := BRISQUE(s^{t}) - BRISQUE(s^{t+1});\\
&r^{t}_i := r^{t,MSE}_i + \lambda r^{t,BRISQUE}_i,\label{eq:5}
\end{align}

where \(BRISQUE(s^{t})\) returns the BRISQUE score of \(s^{t}\) and $\lambda$ is a hyperparameter that balances between the two rewards. Since producing the pseudo-derained reference $y^{pr,t}$ is a stochastic process, the reference must be randomly sampled at each time step.

\begin{figure*}[t!]
\centering
\includegraphics[width=2\columnwidth]{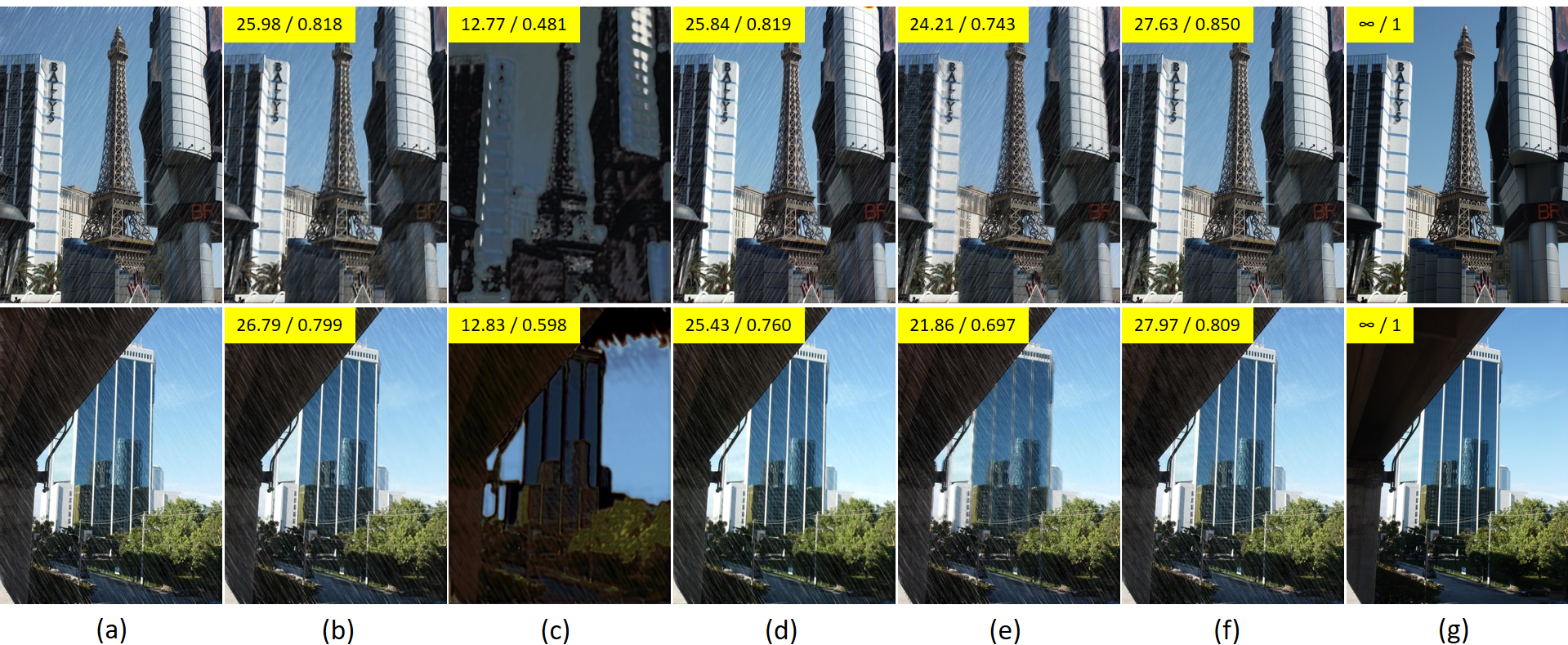}
\vspace{-1em}
\caption{Qualitative comparisons on DDN-SIRR\_syn with PSNR/SSIM values shown on the results. (a) Input images, and the derained results obtained using (b) DIP~\cite{ulyanov2018deep}, (c) N2S~\cite{batson2019noise2self}, (d) N2V~\cite{krull2019noise2void}, (e) R2A~\cite{peng2023rain2avoid}, and (f) Ours. (g) GT images.}
\label{fig: Quality_DDN_syn}
\vspace{-1.5em}
\end{figure*}

\section{EXPERIMENTS}
\label{sec:exp}

\noindent\textbf{Datasets}: To evaluate the deraining performance, we use four public deraining benchmark datasets, consisting of Rain100L~\cite{yang2017deep},
Rain800~\cite{zhang2019image}, and DDN-SIRR~\cite{wei2019semi}. Rain100L and Rain800 are synthetic datasets, and DDN-SIRR~\cite{wei2019semi} contains $14,000$ synthetic images and $147$ real-world rain images. Here, we use the Rain100L test set, Rain800 test set, and all real-world DDN-SIRR images (dubbed with DDN-SIRR\_real) and $400$ images randomly sampled from the synthetic DDN-SIRR set (dubbed with DDN-SIRR\_syn) as practiced in~\cite{rai2022fluid}. 

\begin{table}[t!]
\centering
\caption{The inpainting action set for SRL-Derain as used in~\cite{furuta2019fully}}
\resizebox{\linewidth}{!}{
\begin{tabular}{ |c|c|c|c|  }
 \hline
 Index & Action & Filter Size & Parameter\\
 \hline
 1 & box filter & $5 \times 5$ & -\\
 \hline
 2 & bilateral filter & $5 \times 5$ & $\sigma_c=1.0, \sigma_s=5$\\
 \hline
 3 & bilateral filter & $5 \times 5$ & $\sigma_c=0.1, \sigma_s=5$\\
 \hline
 4 & median filter & $5 \times 5$ & -\\
 \hline
 5 & gaussian filter & $5 \times 5$ & $\sigma=1.5$\\
 \hline
 6 & gaussian filter & $5 \times 5$ & $\sigma=0.5$\\
 \hline
 7 & pixel value += 1 & - & - \\
 \hline
 8 & pixel value -= 1 & - & - \\
 \hline
 9 & do nothing & - & - \\
 \hline
\end{tabular}
}
\vspace{-1.7em}
\label{tab:action_set}  
\end{table}

\noindent\textbf{Implementation Details.} We train the RL agents using the Adam optimizer for $100$ episodes with the initial learning rate starting from $1 \times 10^{-3}$ and multiplied by $(1-\frac{episode}{max\_episode})^{0.9}$ at each episode, where $max\_episode$ is set to $150$. The length of each episode \(t_{max}\) is set to 15. The action set is listed in Table~\ref{tab:action_set}. The \(\lambda\) in Eq.~\ref{eq:5} is set to $0.025$ empirically. The model is implemented in Python with Chainer and trained on a computer equipped with an Intel Xeon Silver 4210 CPU and a single ASUS TURBO RTX 3090 GPU with 24G memory.

\noindent\textbf{Image Quality Metrics and Compared Methods.} For synthetic data with the Ground Truth (GT), we evaluate the performance with PSNR and SSIM. On the other hand, we use no-reference image quality metrics, the Blind/Referenceless Image Spatial Quality Evaluator (BRISQUE)~\cite{mittal2012no}, for real-world rain images without their GT. 
BRISQUE is a score calculated based on the extracted Natural Scene Statistics.
A larger value for PSNR or SSIM means better performance, while a smaller value for BRISQUE indicates better quality.
We compare our SRL-Derain with state-of-the-art deraining and denoising approaches, including a dictionary-learning-based deraining method (Kang's~\cite{decompositionSID_Kang2011TIP}), a prior-based conventional deraining method (David's~\cite{eigen2013restoring}), a semi-supervised deraining method (Wei's~\cite{wei2019semi}), a few-shot self-supervised deraining method (FLUID~\cite{rai2022fluid}), three self-supervised denoising methods (DIP~\cite{ulyanov2018deep}, N2V~\cite{krull2019noise2void}, and N2S~\cite{batson2019noise2self}), and self-supervised deraining method (R2A~\cite{peng2023rain2avoid}).

\begin{figure*}[t!]
\centering
\includegraphics[width=2\columnwidth]{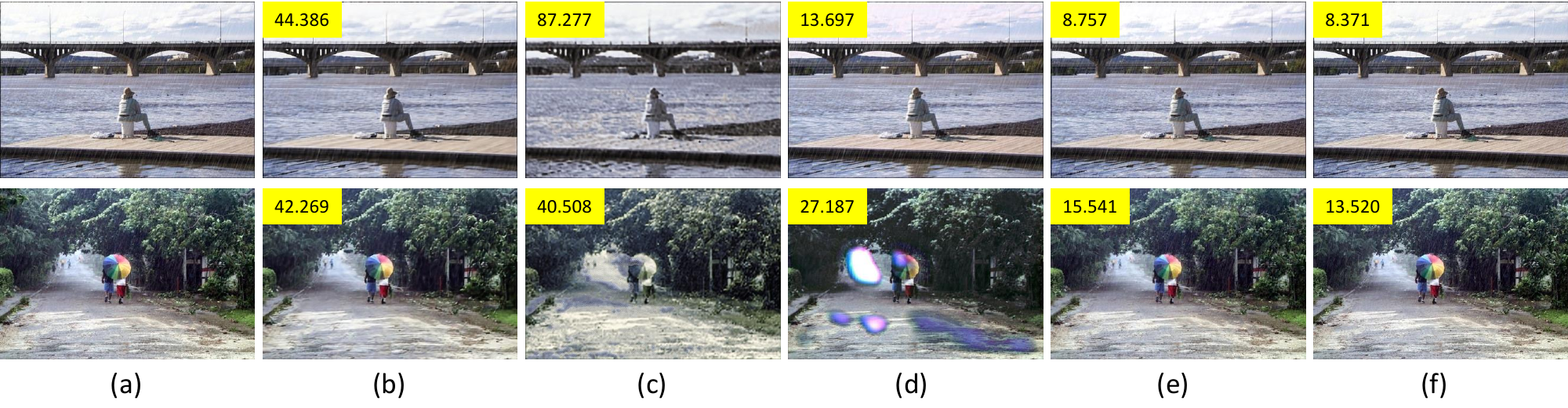}
\vspace{-1em}
\caption{Qualitative comparisons on DDN-SIRR\_real with BRISQUE values shown on the results. (a) Input images, and the derained results obtained using (b) DIP~\cite{ulyanov2018deep}, (c) N2S~\cite{batson2019noise2self}, (d) N2V~\cite{krull2019noise2void}, (e) R2A~\cite{peng2023rain2avoid}, and (f) Ours.}
\label{fig: Quality_DDN_real}
\vspace{-1em}
\end{figure*}

\begin{table}[t!]
\centering
\caption{Evaluation results on Rain800, DDN-SIRR\_syn and DDN-SIRR\_real. The best scores are in bold, and the second best is underlined.}
\resizebox{\columnwidth}{!}{
\begin{tabular}{|c|c|c|c|c|c|}
\hline
\multicolumn{1}{|c|}{\multirow{2}*{Method}}&\multicolumn{2}{c|}{Rain800}&\multicolumn{2}{c|}{DDN-SIRR\_syn}&\multicolumn{1}{c|}{DDN-SIRR\_real}\\
\cline{2-6}
\multicolumn{1}{|c|}{}&PSNR$\uparrow$&SSIM$\uparrow$&PSNR$\uparrow$&SSIM$\uparrow$&BRISQUE$\downarrow$\\
\hline
\multirow{1}{*}{Kang's~\cite{decompositionSID_Kang2011TIP}}&21.20&\underline{0.730}&22.65&0.712&122.063\\
\hline
\multirow{1}{*}{David's~\cite{eigen2013restoring}}&18.95&0.663&19.18&0.681&38.581\\
\hline
\multirow{1}{*}{DIP~\cite{ulyanov2018deep}}&22.24&0.684&23.49&0.725&60.540\\
\hline
\multirow{1}{*}{N2S ~\cite{batson2019noise2self}}&19.11&0.506&19.06&0.496&89.041\\
\hline
\multirow{1}{*}{N2V~\cite{krull2019noise2void}}&21.75&0.659&22.81&0.712&39.702\\
\hline
\multirow{1}{*}{R2A~\cite{peng2023rain2avoid}}&\underline{22.76}&\textbf{0.732}&\underline{24.61}&\underline{0.784}&\underline{30.141}\\
\hline
\multirow{1}{*}{SRL-Derain (Ours)} & \textbf{23.13} & 0.723 & \textbf{24.91} & \textbf{0.789} & \textbf{29.308}\\
\hline
\end{tabular}
}
\vspace{-1em}
\label{tab:objsyn}
\end{table}

\begin{table*}[t!]
\centering
\caption{Evaluation results on Rain100L. Note that the results
of Wei's~\cite{wei2019semi} and FLUID (1-, 3-, and 5-shot) are directly excerpted from~\cite{rai2022fluid}.}
\begin{adjustbox}{width=2\columnwidth}
\begin{tabular}{|c|c|c|c|c|c|c|c|c|c|c|c|c|c|}
\hline
{Rain100L}&{Kang's}&{David's}&{DIP}&{N2S}&{N2V}&{Wei's}&{FLUID (1)}&{FLUID (3)}&{FLUID (5)}&{R2A}&{SRL-Derain (Ours)}\\
\cline{1-12}
PSNR$\uparrow$&24.46&19.94&24.98&20.20&21.74&23.77&23.87&25.54&{26.87}&\underline{29.17}&\textbf{29.86}\\
\hline
SSIM$\uparrow$&0.713&0.744&0.704&0.498&0.648&0.775&0.772&0.826&0.862&\underline{0.887}&\textbf{0.906}\\
\hline
\end{tabular}
\end{adjustbox}
\vspace{-1.5em}
\label{tab:objrain100}
\end{table*}

\noindent\textbf{Quantitative Analysis.} 
We compare with existing methods~\cite{decompositionSID_Kang2011TIP, eigen2013restoring, ulyanov2018deep, batson2019noise2self, krull2019noise2void, peng2023rain2avoid}, on the Rain800, DDN-SIRR\_syn, and DDN-SIRR\_real datasets, as presented in Table~\ref{tab:objsyn}.  Our method performs favorably against the compared methods on these datasets. Additionally, we compare our approach with~\cite{decompositionSID_Kang2011TIP, eigen2013restoring,ulyanov2018deep,batson2019noise2self,krull2019noise2void,peng2023rain2avoid}, the semi-supervised deraining method Wei's~\cite{wei2019semi}, and the few-shot self-supervised deraining method FLUID~\cite{rai2022fluid} with $1$-shot, $3$-shot, and $5$-shot settings on Rain100L. Table~\ref{tab:objrain100} shows our SRL-Derain achieves the best results, even better than FLUID~\cite{rai2022fluid} with $3$-shot or $5$-shot settings ($3$ or $5$ more training images used).

\noindent\textbf{Qualitative Analysis.}
We also compare the derained results of our method with those using the compared self-supervised denoising methods~\cite{ulyanov2018deep, batson2019noise2self, krull2019noise2void} and self-supervised deraining approach~\cite{peng2023rain2avoid}. Figure~\ref{fig: Quality_rain100L} shows some examples of deraining results on the Rain100L testing set, where DIP~\cite{ulyanov2018deep} seems only to blur the images. N2S~\cite{batson2019noise2self} removes most rain streaks but causes color distortions. N2V~\cite{krull2019noise2void} removes partial rain but introduces color artifacts. R2A~\cite{peng2023rain2avoid} also removes most rain, whereas the proposed SRL-Derain works the best.
Figure \ref{fig: Quality_rain800} and Figure ~\ref{fig: Quality_DDN_syn} demonstrate two derained results on the Rain800 testing set and the DDN-SIRR\_syn testing set. As can be seen, we have similar observations to those for Figure~\ref{fig: Quality_rain100L}, where SRL-Derain also performs favorably. The PSNR and SSIM values shown in the results can also validate these visualizations.
Figure~\ref{fig: Quality_DDN_real} shows the results on the DDN-SIRR\_real dataset, where the abovementioned observations for the synthetic rain images can also apply. We can find that R2A and the proposed SRL-Derain work comparably, both of which restore the rain images nicely.

\noindent\textbf{Ablation Study and Analysis.} 
In this section, we conduct an ablation study on two designs we adopt to verify their effectiveness. First, to be fair in comparison with self-supervised denoising methods, which do not have a priori knowledge about rain like R2A~\cite{peng2023rain2avoid} (w/ LDGP) and our SRL-Derain (w/ RDP), we provide those methods with the Ground Truth (GT) rain mask for them to not denoise non-rain pixels, which may cause unwanted blur. Note that the study uses the Rain100L training set since it has the GT rain masks. Table~\ref{tab:rain_mask} shows the self-supervised denoising methods (N2V and DIP) using the GT rain mask to avoid smoothing non-rain pixels perform better than without the mask. R2A with the GT mask also works better than its original LDGP. As also can be seen, our SRL-Derain using LDGP achieves higher PSNR/SSIM values than R2A, showing our proposed self-supervised RL scheme outperforms the SSL deraining method. SRL-Derain using the GT mask works better than the other methods with the setting. 
In addition, Table~\ref{tab:reward} verifies the effectiveness of adding the no-reference quality metric, BRISQUE, to the reward function on the Rain100L testing set, where SRL-Derain using BRISQUE as a reward can indeed improve the BRISQUE score.
At last, we can train SRL-Derain on multiple images in an SSL manner. In our experiment, SRL-Derain can achieve $29.93$dB in PSNR, higher than R2A~\cite{peng2023rain2avoid} by $0.88$dB, and our average inference time for running $15$ steps is $1.05$ seconds. 
\begin{table}[t!]
\vspace{-1em}
\caption{Ablation on effects of the rain mask provided for self-supervised methods.}
\vspace{-0.2em}
\label{tab:rain_mask}
\begin{center}
\begin{adjustbox}{width=0.7\columnwidth}
\begin{tabular}{|p{4cm}||p{1.2cm}|p{1.2cm}|}
 \hline
 \multicolumn{3}{|c|}{Rain100L\_train} \\
 \hline
 Rain Mask&PSNR$\uparrow$&SSIM$\uparrow$\\
 \hline
 N2V~\cite{krull2019noise2void} & 22.74 & 0.654 \\
 N2V w/ GT Mask & 25.99 & 0.822 \\
 DIP~\cite{ulyanov2018deep} & 24.79 & 0.709 \\
 DIP w/ GT Mask & 27.76 & 0.781 \\
 R2A w/ LDGP~\cite{peng2023rain2avoid} & 28.79 & 0.895 \\
 R2A w/ GT Mask & \underline{31.08} & \underline{0.927} \\
 SRL-Derain w/ LDGP & 29.09 & 0.895 \\
 SRL-Derain w/ RDP & 29.82 & 0.906 \\
 SRL-Derain w/ GT Mask & \textbf{32.50} & \textbf{0.929} \\
 \hline
\end{tabular}
\end{adjustbox}
\vspace{-2em}
\end{center}
\end{table}

\section{Conclusions} 
This paper presents SRL-Derain, a self-supervised RL scheme for image deraining. By incorporating pseudo-derained references and a no-reference image quality metric as self-supervised rewards, we can successfully train an RL model to progressively derain the input rain image. The experimental results indicated that the proposed method performed favorably against the SOTA few-shot deraining and self-supervised denoising and deraining methods. We hope this work can inspire future research on using self-supervised RL for low-level vision tasks.

\begin{table}[h!]
\vspace{-1.2em}
\caption{Ablation on the BRISQUE reward in the proposed SRL-Derain.}
\vspace{-.5em}
\label{tab:reward}
\begin{center}
\begin{adjustbox}{width=.8\columnwidth}
\begin{tabular}{|p{3cm}||p{1.2cm}|p{1.2cm}|p{1.6cm}|}
 \hline
 \multicolumn{4}{|c|}{Rain100L\_test} \\
 \hline
 Reward &PSNR$\uparrow$&SSIM$\uparrow$&BRISQUE$\downarrow$\\
 \hline
 w/o BRISQUE & \textbf{29.86} & 0.904 & 11.891 \\
 w/ BRISQUE & \textbf{29.86} & \textbf{0.906} & \textbf{10.071} \\
 \hline
\end{tabular}
\end{adjustbox}
\vspace{-2em}
\end{center}
\end{table}

\bibliographystyle{IEEEbib}
\bibliography{ref}

\end{document}